\PassOptionsToPackage{table,dvipsnames}{xcolor}
\documentclass[sigconf, natbib=false]{acmart}
\AtBeginDocument{%
  }

\setcopyright{acmlicensed}
\copyrightyear{2025}
\acmYear{2025}
\acmDOI{XXXXXXX.XXXXXXX}
\acmConference[KDD '25] {Proceedings of the 31st ACM SIGKDD Conference on Knowledge Discovery and Data Mining V.2}{August 3--7, 2025}{Toronto, ON, Canada.}
\acmISBN{979-8-4007-1454-2/25/08}
\acmDOI{10.1145/3711896.3737271}

\acmSubmissionID{adsb0222}

\RequirePackage[
  datamodel=acmdatamodel,
  style=acmnumeric,
  ]{biblatex}

\usepackage{multirow}
\usepackage{subcaption}
\usepackage{graphicx}
\usepackage[super]{nth}
\usepackage[normalem]{ulem}
\useunder{\uline}{\ul}{}

\usepackage{listings}
\usepackage{xspace}

\newcommand{\MLSNT}{MLSNT}

\begin{document}

\title[Unified Game Moderation]{Unified Game Moderation: Soft-Prompting and LLM-Assisted Label Transfer for Resource-Efficient Toxicity Detection}

\author{Zachary Yang}
\email{zachary.yang@mail.mcgill.ca}
\affiliation{%
  \institution{Ubisoft La Forge | McGill University |  Mila}
  \city{Montreal}
  \country{Canada}
}

\author{Domenico Tullo}
\affiliation{%
  \institution{Ubisoft La Forge}
  \city{Montreal}
  \country{Canada}
}

\author{Reihaneh Rabbany}
\affiliation{%
  \institution{McGill University |  Mila | CIFAR AI Chair}
  \city{Montreal}
  \country{Canada}
}

\renewcommand{\shortauthors}{Yang et al.}

\begin{abstract}

Toxicity detection in gaming communities faces significant scaling challenges when expanding across multiple games and languages, particularly in real-time environments where computational efficiency is crucial. We present two key findings to address these challenges while building upon our previous work on ToxBuster, a BERT-based real-time toxicity detection system. First, we introduce a soft-prompting approach that enables a single model to effectively handle multiple games by incorporating game-context tokens, matching the performance of more complex methods like curriculum learning while offering superior scalability. Second, we develop an LLM-assisted label transfer framework using GPT-4o-mini to extend support to seven additional languages.
Evaluations on real game chat data across French, German, Portuguese, and Russian achieve macro F1-scores ranging from 32.96\% to 58.88\%, with particularly strong performance in German, surpassing the English benchmark of 45.39\%. In production, this unified approach significantly reduces computational resources and maintenance overhead compared to maintaining separate models for each game and language combination. At Ubisoft, this model successfully identifies an average of 50 players, per game, per day engaging in sanctionable behavior.

\end{abstract}



\received{10 February 2025}
\received[accepted]{22 May 2025}

\maketitle
\section{Introduction}

Online toxicity has become a pervasive challenge across digital platforms, with gaming communities facing particularly severe impacts. The Anti-Defamation League (ADL) reports alarming harassment rates of 76\% among adults and 75\% among teens and pre-teens in gaming spaces as of 2023 \cite{adl_2023}. Beyond the documented psychological harm and potential to incite real-world violence \cite{adl_2021}, toxic behavior directly impacts gaming companies' revenue streams, with 20\% of players reducing their spending due to harassment experiences. This challenge spans multiple platforms, from social media (Facebook \cite{hate_speech_on_facebook}, Reddit \cite{impact_of_toxic_language_on_reddit}, YouTube \cite{hate_speech_in_youtube}) to various gaming environments \cite{playing_against_hate_speech, yang-etal-2023-towards-detecting}.

To address this challenge, our previous work introduced ToxBuster \cite{yang-etal-2023-towards-detecting}, a real-time chat toxicity detection model designed for production deployment. However, scaling this solution across Ubisoft's diverse game portfolio revealed two critical challenges: the need to unify game-specific models into a single deployable solution and the requirement to support multiple languages. These challenges reflect broader industry needs for scalable, multi-lingual content moderation systems.

The first challenge emerges from the operational complexity of maintaining separate models for each game. While our initial approach of creating game-specific models demonstrated effectiveness, it proved unsustainable as our game portfolio expanded. After exploring several approaches to unification, including mixed dataset training and curriculum learning, we successfully adapted soft-prompting techniques to maintain performance while significantly improving scalability.

The second challenge involves extending toxicity detection across languages while addressing the fundamental issue of inconsistent toxicity definitions. Despite growing research attention, the field lacks standardized definitions \cite{https://doi.org/10.48550/arxiv.1809.07572}, leading to varied approaches in task formulation \cite{Vidgen2020}. This challenge is particularly acute when organizations need to update their toxicity categories over time, as we experienced at Ubisoft. To address this, we developed an innovative LLM-assisted label transfer framework that leverages existing datasets while adapting to organization-specific definitions. We applied this framework to 15 open-source toxicity datasets spanning Chinese, French, German, Japanese, Portuguese, and Russian, creating a comprehensive Multi-Lingual Social Network Toxicity (\MLSNT) dataset aligned with our current toxicity definitions.

In summary, our contributions are the following:
\begin{enumerate}
\item A soft-prompting approach that successfully unifies game-specific toxicity models while maintaining performance and improving scalability
\item A new multi-lingual toxicity dataset, \MLSNT,  created through our LLM-assisted label transfer framework
\item A unified, production-ready toxicity detection model that effectively handles multiple games and languages
\end{enumerate}

\textbf{Reproducibility:} The dataset is released \href{https://huggingface.co/datasets/ComplexDataLab/MLSNT}{ComplexDataLab/MLSNT}.

\section{Related Works}

Toxicity detection in online platforms has garnered significant attention over the past decade. Early work framed toxicity detection as a straightforward classification task using traditional machine learning methods with hand-crafted features \cite{hate_speech_on_twitter}. Subsequent approaches leveraged deep neural networks to better capture context and nuance in text \cite{gamback-sikdar-2017-using, content_driven_detection}, while more recent studies have embraced pre-trained language models to exploit their contextualized representations \cite{ALMEREKHI2022100019, DBLP:journals/corr/abs-2201-00598, perspective_api}. In online gaming environments, where rapid, real-time interactions necessitate quick and context-aware responses, researchers have developed specialized detection systems \cite{10.1145/2998181.2998213, yang-etal-2023-towards-detecting, 10.1145/3675805}. The global nature of online communication has further spurred research into multi-modal and multi-lingual toxicity detection, with works integrating cross-lingual transfer, visual cues, and audio signals to robustly address toxicity across diverse communities \cite{zampieri-etal-2019-semeval, bui2024multi3hatemultimodalmultilingualmulticultural, cao-etal-2024-toxicity}.

Scaling language models (for toxicity detection) to handle large, heterogeneous datasets has necessitated the exploration of varied training paradigms. While early approaches employed end-to-end training on single-task objectives, recent work has highlighted the benefits of mixed training regimes that combine supervised fine-tuning with unsupervised pre-training \cite{kaplan2020scalinglawsneurallanguage, brown2020languagemodelsfewshotlearners, ke2023continualpretraininglanguagemodels}. Our work incorporates continuous pre-training and domain adaptation of the base model \cite{farahani2020briefreviewdomainadaptation}. Curriculum learning strategies, where models are gradually exposed to increasingly challenging examples, have proven effective in managing class imbalance and contextual variability \cite{10.1145/1553374.1553380, soviany2022curriculumlearningsurvey}. We implement a naive version of this approach. Complementary advances in soft-prompting and meta-learning have enabled more efficient fine-tuning of large pre-trained models, reducing computational overhead while maintaining strong performance on toxicity detection tasks \cite{huang-etal-2023-learning-better, li2021prefixtuningoptimizingcontinuousprompts}.

The emergence of large language models has expanded toxicity detection into the realm of zero-shot and few-shot learning. Pioneering studies have demonstrated that models like GPT-3 \cite{brown2020languagemodelsfewshotlearners} can generalize to detect harmful content with minimal task-specific supervision. This has paved the way for prompt-based approaches that elicit structured reasoning for improved identification of toxic content \cite{koh2024llmsrecognizetoxicitystructured, shaikh-etal-2023-second, PAN20242849, plaza-del-arco-etal-2023-respectful}. While these works underscore the versatility of LLMs in handling toxicity detection, challenges such as bias, underperformance compared to fine-tuned smaller language models in benchmarks, and interpretability continue to stimulate further research, as highlighted in recent surveys \cite{bommasani2022opportunitiesrisksfoundationmodels, cao-etal-2024-toxicity}.

Real-time toxicity detection remains critical for ensuring safe online interactions, particularly in dynamic settings like social media and gaming. Small language models (SLMs) are still the most practical solution, striking an attractive balance between the upfront cost of dataset creation and detection performance, inference speed, and cost efficiency \cite{yu2023openclosedsmalllanguage} compared to the continuous cost of large language models. These models benefit from streamlined architectures and efficient inference techniques essential for deployment in resource-constrained environments. Recent empirical studies validate that SLMs can moderate content in real time without sacrificing accuracy, thereby offering practical advantages over heavier, more computationally expensive alternatives \cite{yu2023openclosedsmalllanguage, zhang2023efficienttoxiccontentdetection}.
\section{Methodology}

Our methodology addresses two key production challenges: unifying game-specific models into a single deployable model and extending toxicity detection capabilities to multiple languages. This section details our approaches to both challenges.

\subsection{Unifying Game-Specific Models}
Production deployment requirements necessitated consolidating our game-specific toxicity detection models into a single, unified model capable of handling multiple games. Previously, ToxBuster used separate models for each game (two popular multi-player games), which increased operational complexity and resource usage. We investigated approaches to maintain model effectiveness while simplifying the deployment architecture, evaluating four training approaches of increasing complexity:
\begin{enumerate}
    \item \textbf{Single-game (baseline):} Individual models fine-tuned on game-specific datasets, representing our previous work. In production, we would select the highest performing model.
    \item \textbf{Mixed-dataset:} A unified approach combining data from all games into a single dataset for fine-tuning one model, significantly simplifying deployment.
    \item \textbf{Curriculum learning:} An extension of the mixed-dataset approach where we continue fine-tuning on individual game datasets sequentially. While this produces multiple intermediate models, we select the best-performing model across all games for deployment.
    \item \textbf{Soft prompting:} An approach that maintains a single model while preserving game-specific knowledge through specialized input tokens. We prepend a GAME\_TYPE\_TOKEN to each input sequence, with three possible values: GAME\_1, GAME\_2, or GAME\_UNKNOWN. This allows the model to adapt its behavior based on the gaming context. The GAME\_UNKNOWN token, while not trained, can be used during inference when the game is unknown (due to API/production constraints) or when testing a new game.
\end{enumerate}

All experiments fine-tunes ToxBuster -- using bert-base-uncased with a maximum length of 512 tokens, incorporating chat context up to this limit. To ensure robust evaluation, we conducted five runs with different random seeds. For soft prompting experiments, we evaluated both game-aware (GAME\_x tokens) and game-agnostic (GAME\_UNKNOWN token) scenarios to simulate real-world deployment conditions. Performance was measured using macro F1-score for each game dataset and the overall average across games.

\subsection{Multi-lingual Extension}

\begin{figure*}[htbp!]
    \centering
    \includegraphics[scale=0.6]{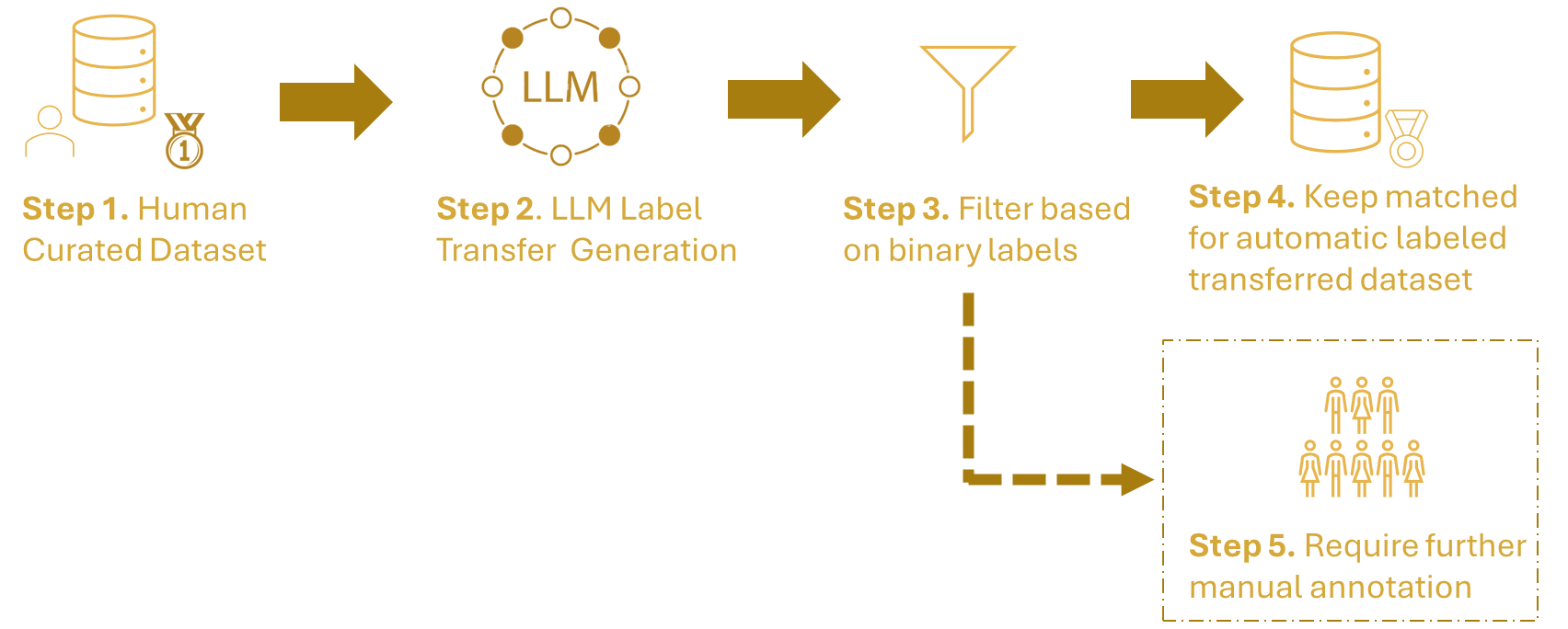}    
    \caption{LLM-assisted Label Transfer Framework for Cross-dataset Toxic Content Classification. The framework leverages existing human-curated datasets through LLM-based label generation, followed by a validation filter that ensures consistency between human and LLM annotations.}
    \Description{LLM-assisted Label Transfer Framework for Cross-dataset Toxic Content Classification. The framework leverages existing human-curated datasets through LLM-based label generation, followed by a validation filter that ensures consistency between human and LLM annotations.}
    \label{fig:llm-assisted-label-transfer}
\end{figure*}

\subsubsection{Base Model Selection}
To extend ToxBuster's capabilities beyond English, we evaluated several pre-trained multi-lingual BERT variants as replacements for our existing BERT-based model. We compared four architectures and their multi-lingual counterparts (BERT, DistilBERT, DeBERTa, and RoBERTa) using Game\_1 data as our benchmark. XLM-RoBERTa-base emerged as our optimal choice based on three criteria: superior macro F1-score performance, simplified continuous pre-training process (requiring only masked language modeling instead of both MLM and next sentence prediction), and its pre-training on the comprehensive CC-100 dataset.

\subsubsection{Language Identification}
We analyzed the language distribution in our game chat data using Lingua-py\footnote{\url{https://github.com/pemistahl/lingua-py}} to prioritize our multi-lingual expansion. While this tool's accuracy decreases with shorter text lengths—a limitation relevant to typically brief game chat messages—it provided sufficient insight for our initial language assessment. We analyzed one week of chat data from both games focusing on 12 languages that intersect with CC-100 dataset coverage, Lingua's capabilities, and our global player base: English, Spanish, French, German, Russian, Portuguese, Japanese, Korean, Chinese, Arabic, Hindi, and Thai. Our analysis revealed that English dominates the chat data at 80\%, followed by French, German, Japanese, Portuguese, and Russian as the next most frequent languages.

\subsubsection{LLM-assisted Label Transfer}

A fundamental challenge in toxicity detection is the inconsistency of definitions across datasets. Even within a single organization, toxic categories evolve as business rules and legislation change, leading to definitions being updated, split, or merged over time. While numerous existing datasets could potentially be leveraged, obtaining expert review for label standardization is challenging and costly. To address this, we propose an LLM-assisted label transfer approach.
Our approach is based on three key assumptions:
\begin{enumerate}
    \item The original labels in existing datasets (both proprietary and open-source) are considered ground truth, having been human-annotated.
    \item There is sufficient agreement in binary classification across datasets—content labeled as toxic or non-toxic in source datasets aligns with our use case definitions.
    \item For fine-grained toxic categories, we can accept lower performance (e.g., accuracy, F1-score).
\end{enumerate}

We implement the label transfer through the following workflow, as illustrated in Figure \ref{fig:llm-assisted-label-transfer}:

\begin{enumerate}
    \item Source a human-annotated dataset related to toxicity detection
    \item Convert the original human annotations to binary classifications
    \item Use an LLM (e.g., GPT-4o-mini) to generate both binary (toxic/non-toxic) and specific category labels
    \item Retain only the entries where human and LLM binary classifications agree
\end{enumerate}

We evaluated GPT-4o-mini's ability to align with our toxicity classification framework using a prompt that combines meta prompting \cite{zhang2024metapromptingaisystems}, chain of thought \cite{wei2023chainofthoughtpromptingelicitsreasoning}, and self consistency \cite{wang2023selfconsistencyimproveschainthought}. The model was tasked with providing binary classification (toxic/non-toxic), identifying toxic spans, and assigning specific toxicity categories. We validated this approach using our golden dataset, which consists of entries with unanimous agreement among three annotators.
Due to certain constraints, our current implementation ends here. Our planned next step would be having moderators review the discarded cases where the previous human annotators and LLM labels diverge. The key advantage of this approach is its ability to efficiently collect clear-cut toxic and non-toxic cases in non-English languages, reserving valuable human annotation resources for ambiguous cases requiring nuanced judgment.

\subsubsection{Human-Annotated Datasets}
We collected human-annotated hate speech datasets from \url{hatespeechdata.com} and other peer-reviewed publications, covering seven languages: French (1 dataset \cite{ousidhoum-etal-2019-multilingual}), German (3 datasets \cite{goldzycher-etal-2024-improving, germ_eval, 10.1145/3368567.3368584}), Portuguese (3 datasets \cite{brasnam,10.1007/s10579-023-09657-0,leite-etal-2020-toxic}), Russian (2 datasets \cite{DBLP:conf/raslan/AndrusyakRK18, saitov-derczynski-2021-abusive}), Simplified Chinese (3 datasets \cite{deng-etal-2022-cold,JIANG2022100182,lu-etal-2023-facilitating}), Traditional Chinese (1 dataset \cite{10.1109/IRI51335.2021.00069}), and Japanese (2 datasets \cite{DBLP:journals/corr/abs-2407-03963}). Table \ref{tab:dataset_overivew} details each dataset's original classification task, source platform, data volume, and estimated computational cost for GPT-4o-mini-assisted label transfer. We applied our LLM-assisted label transfer framework to convert these 15 datasets into our toxicity categories, forming the Social Network System (SNS) dataset.

\subsubsection{ToxBuster}

We adapted ToxBuster for multi-lingual capability by evaluating three base models: bert-base (baseline), xlm-roberta-base, and xlm-roberta-base-adapted. The xlm-roberta-base-adapted model underwent pre-training on a week's worth of chat data from both games for 74 epochs, with early stopping when masked language model loss plateaued (maximum 100 epochs). We applied soft-prompting with GAME\_TYPE\_TOKEN values: GAME\_1, GAME\_2, \MLSNT, and GAME\_UNKNOWN. Performance was measured using weighted and macro F1-scores.

\subsubsection{Human-eval}

To evaluate \MLSNT's transferability to both games, we developed a systematic sampling strategy for each game and language. We targeted 50 samples per category with a 20\% spillover mechanism to the next toxic category and 80\% to the non-toxic category, maintaining toxic imbalances similar to English. By dynamically calculating available samples and adjusting targets based on shortfalls, the approach ensures a more balanced representation of toxic and non-toxic content. One human annotator proficient in each language and familiar with the games validated their respective evaluation set of 450 lines.

\section{Results}

We present our experimental results in two parts: first, we evaluate our approach to scale a single model across multiple games, followed by our results on extending the model to support seven languages.

\subsection{Unifying Game-Specific Models}

\begin{table}[htbp!]
\centering
\caption{Macro F1-score for single, mixed, curriculum learning and soft prompting. For the single-game, we train on the same dataset as the test set. For the rest, we show the training dataset in the brackets, where ``\&'' signifies the mixing of the dataset and ``|'' separates the steps in training. ** Here, we evaluate soft prompting where the game is not provided during inference. Soft prompting achieves comparable performance to curriculum learning even without explicit game information during inference, demonstrating its effectiveness for cross-game generalization.}
\label{tab:single_mixed_cirriculum_soft-prompting_results}
\resizebox{\linewidth}{!}{
\begin{tabular}{cl|ll|l}
\toprule
\multicolumn{2}{c|}{\textbf{Method}} & \textbf{G\_1} & \textbf{G\_2} & {\textbf{Overall}} \\ \midrule
\multicolumn{1}{c}{Single} & (Game\_1) or (Game\_2) & 42.84$\pm$2.56 & 38.80$\pm$0.53 & 40.82$\pm$0.87 \\ \hline
\multicolumn{1}{c}{Mixed} & (G\_1 \& G\_2) & 45.10$\pm$0.73 & 39.87$\pm$2.09 & 42.48$\pm$0.74 \\ \hline
\multirow{2}{*}{Curriculum} & (G\_1 \& G\_2 | G\_1) & 46.32$\pm$0.40 & 40.37$\pm$2.12 & 43.35$\pm$1.16 \\
 & (G\_1 \& G\_2 | G\_2) & 44.19$\pm$1.82 & 40.23$\pm$2.18 & 42.21$\pm$1.52 \\ \hline
 \multirow{2}{*}{\begin{tabular}[c]{@{}c@{}}Soft\\ Prompting \end{tabular}} & (G\_1 \& G\_2)  & 45.39$\pm$1.01 & 40.94$\pm$1.38 & \cellcolor{ACMBlue!8}{43.16$\pm$1.06} \\ 
  &(G\_1 \& G\_2)** & 45.24$\pm$1.13 & 40.82$\pm$1.46 & \cellcolor{ACMBlue!8}{43.03$\pm$1.19} \\ 
 \bottomrule
\end{tabular}
}
\end{table}

In Table \ref{tab:single_mixed_cirriculum_soft-prompting_results}, we compare four different approaches using macro token-level F1-scores on GAME\_1 and GAME\_2 datasets. The overall score represents the balanced F1-score across both games, averaged over five runs. 

Single-game models, as the baseline, achieved a macro F1-score of 42.84\% for GAME\_1 and 38.80\% for GAME\_2. Combining datasets through mixed training improved overall performance to 42.48\%, revealing that the mixed datasets actually benefits both games (+1.7\%). Curriculum learning achieved the highest macro F1-score of 43.35\% with the (GAME\_1 \& GAME\_2 | GAME\_1) sequence, but this approach suffers from significant scalability constraints —adding each new game requires training on the combined dataset of all games, followed by training on the game-specific dataset, and extensive model selection to optimize overall performance. 

Soft-prompting emerges as a more practical solution, achieving the second-highest macro F1-score when the game type is provided during inference and third-highest when it is not. This approach offers superior scalability and efficiency advantages:
\begin{enumerate}
    \item Training is required only once on the combined multi-game dataset, substantially decreasing training time compared to curriculum learning
    \item Supporting new games requires merely introducing a new GAME\_TYPE\_TOKEN and retraining.
\end{enumerate}

\subsubsection{Ablation}

\begin{table}[htbp!]
\centering
\caption{Impact of GAME\_TYPE\_TOKEN position for soft-prompting. GAME\_TYPE\_TOKEN needs to be before the CONTEXT section.}
\label{tab:soft_prompting_ablation}
\resizebox{\linewidth}{!}{
\begin{tabular}{cl|l}
\toprule
\multicolumn{2}{c|}{\textbf{Method}} & \textbf{Overall} \\ \midrule
\multicolumn{2}{c|}{Mixed} & 42.48$\pm$0.74 \\ \hline
\multirow{2}{*}{\begin{tabular}[c]{@{}c@{}}Soft   Prompting\\      (Before Context)\end{tabular}} & (GAME\_1 \& GAME\_2) & \textbf{43.16$\pm$1.06} \\
 & (GAME\_1 \& GAME\_2) ** & 43.03$\pm$1.19 \\ \hline
\multirow{2}{*}{\begin{tabular}[c]{@{}c@{}}Soft Prompting\\      (Before Current Line)\end{tabular}} & (GAME\_1 \& GAME\_2) & 41.84$\pm$1.32 \\
 & (GAME\_1 \& GAME\_2) ** & 41.57$\pm$1.37 \\ 
 \bottomrule
\end{tabular}}
\end{table}

Given our model's unique architecture, which incorporates separate sections for context and the current line, we conducted an ablation study to determine the optimal placement of the GAME\_TYPE\_TOKEN. We compared its effectiveness when positioned before the context versus immediately preceding the current line. As shown in Table \ref{tab:soft_prompting_ablation}, the GAME\_TYPE\_TOKEN is most effective when placed before the context. Performance actually drops if placed before the current line.

\subsection{Multi-lingual Extension}

\subsubsection{Feasibility of LLM-assisted Label Transfer}

\begin{table}[htbp!]
\centering
\caption{GPT-4o-mini label transfer. * represents the dataset with updated set of toxic categories. LLMs can be used to automate the complex task of transferring labels across different annotation schemas.}
\label{tab:gpt-4o-mini-prompt-temp-search}

\begin{tabular}{lcc|ll|l}
\toprule
\multirow{2}{*}{\textbf{}} & \multicolumn{1}{l}{\multirow{2}{*}{\textbf{Prompt}}} & \multicolumn{1}{l|}{\multirow{2}{*}{\textbf{Temp}}} & \multicolumn{2}{c|}{\textbf{Weighted F1-Score}} & \multicolumn{1}{c}{\multirow{2}{*}{\textbf{\begin{tabular}[c]{@{}c@{}}True\\ Labels\end{tabular}}}} \\
 & \multicolumn{1}{l}{} & \multicolumn{1}{l|}{} & \textbf{Binary} & \textbf{Class-wise} & \multicolumn{1}{c}{} \\ 
\midrule
Full & v1 & 1.0 & 86.66\% & 84.48\% & 136,605 \\ \hline
\multirow{4}{*}{Full*} & v1 & 0.7 & \textbf{85.44\%} & \textbf{82.23\%} & \textbf{133,951} \\
 & v1 & 1.0 & 85.35\% & 81.89\% & 132,009 \\ \cline{2-6} 
 & v2 & 0.7 & 82.51\% & 76.85\% & 127,392 \\
 & v2 & 1.0 & 82.47\% & 76.79\% & 127,305 \\ 
 \bottomrule
\end{tabular}

\end{table}

In Table \ref{tab:gpt-4o-mini-prompt-temp-search}, we evaluate the feasibility of using GPT-4o-mini for label transfer on a full dataset where all three human annotators had complete agreement. The toxic categories were updated with two key modifications: (1) splitting Threats into Life-threatening and Non-life threatening categories, and (2) extracting a specific category for Sexual content/harassment while consolidating other categories under Potentially Toxic.

Experimental results demonstrate the potential of GPT-4o-mini's label transfer capabilities. On the original full dataset, the model achieves a weighted F1-score of 86.66\% for the binary toxic/non-toxic classification task, with a slightly lower weighted class-wise F1-score of 84.48\%. When tested on the dataset with updated toxic categories, the binary weighted F1-score remains consistent, though with a minor decline in class-wise performance.

To optimize label transfer, we investigated two prompt variations (v1 and v2) and temperature settings (0.7 and 1.0). Given our framework's focus on retaining only "true" labels—where both human annotators and the large language model concur—we identified prompt version 1 with a temperature of 0.7 as the optimal configuration for maximizing true label count and weighted F1-score.

\begin{table}[htbp!]
\centering
\caption{GPT-4o-mini label transfer filtered. By restricting to only human-LLM agreed binary labels, we gain ~40\% in performance for toxic categories.}
\label{tab:gpt-4o-mini-filtered-f1-score}
\resizebox{\linewidth}{!}{
\begin{tabular}{lcc|llll}
\toprule
\textbf{} & \multicolumn{1}{l}{\textbf{Prompt}} & \multicolumn{1}{l|}{\textbf{Temp}} & \multicolumn{1}{c}{\textbf{\begin{tabular}[c]{@{}c@{}}No\\ Filter\end{tabular}}} & \multicolumn{1}{c}{\textbf{\begin{tabular}[c]{@{}c@{}}LLM \\ "Toxic"\end{tabular}}} & \multicolumn{1}{c}{\textbf{\begin{tabular}[c]{@{}c@{}}\cellcolor{ACMBlue!8}{Agreed}\\ \cellcolor{ACMBlue!8}{Toxic}\end{tabular}}} & \multicolumn{1}{c}{\textbf{\begin{tabular}[c]{@{}c@{}}Agreed\\ Labels\end{tabular}}} \\ 
\midrule
Full & v1 & 1.0 & 84.48\% & 38.36\% & \cellcolor{ACMBlue!8}{79.12\%} & 98.10\% \\ \hline
\multirow{4}{*}{Full*} & v1 & 0.7 & 82.23\% & 29.30\% & \cellcolor{ACMBlue!8}{67.90\%} & 97.20\% \\
 & v1 & 1.0 & 81.89\% & 29.55\% & \cellcolor{ACMBlue!8}{67.29\%} & 96.97\% \\ \cline{2-7} 
 & v2 & 0.7 & 76.85\% & 22.74\% & \cellcolor{ACMBlue!8}{61.24\%} & 95.66\% \\
 & v2 & 1.0 & 76.79\% & 22.73\% & \cellcolor{ACMBlue!8}{61.41\%} & 95.66\% \\ \bottomrule
\end{tabular}
}
\end{table}

We delve deeper into the weighted class-wise F1-score by examining the filtering approaches. Table \ref{tab:gpt-4o-mini-filtered-f1-score} demonstrates the critical importance of our proposed filtering methodology, which focuses on cases where human annotators and GPT-4o-mini agree. The results reveal a critical insight into our proposed methodology of retaining only cases where human annotators and the large language model agree. When considering unfiltered predictions, the baseline performance is at 84.48\%. However, when we restrict the analysis to GPT-4o-mini's toxic predictions alone, the performance is notably low, reaching only 38.36\%. This underscores the challenges of relying solely on an LLM's classification. The most significant finding emerges when we filter for cases where both human annotators and GPT-4o-mini classify content as toxic: the performance dramatically improves to 79.12\%, representing a remarkable 40\% increase in performance. This consistent pattern holds true across different prompt versions and temperature settings. Hence, our approach offers a solution in minimizing the cost of human annotation when label transfer is needed. Intuitively, can effectively capture unambiguous cases of toxic and non-toxic content, providing a reliable automatic filtering when toxic labels need to be re-assigned.

\subsubsection{Multi-lingual Base Model Search} 

\begin{table}[hbtp!]
\centering
\caption{Multi-lingual base BERT model search. Considering the recency of the training dataset and model performance, xlm-roberta-base is preferred.}
\label{tab:multi-lingual-base-model-search}
\resizebox{\linewidth}{!}{
\begin{tabular}{l|l|ll}
\toprule
\multicolumn{1}{c|}{\multirow{2}{*}{\textbf{Model}}} & \multirow{2}{*}{\textbf{Params}} & \multicolumn{2}{c}{\textbf{F1-Score}} \\
\multicolumn{1}{c|}{} &  & \textbf{Macro} & \textbf{Weighted}  \\ \hline
bert-base-uncased & 110M & 43.30$\pm$1.73 & 82.02$\pm$0.33 \\
\cellcolor{ACMBlue!8}{...-multilingual-cased} & \cellcolor{ACMBlue!8}{179M} & \cellcolor{ACMBlue!8}{39.31$\pm$2.68} & \cellcolor{ACMBlue!8}{81.28$\pm$0.49} \\ 
\hline
distilbert-base-uncased & 67M & 41.07$\pm$2.05 & 81.46$\pm$0.66 \\
\cellcolor{ACMBlue!8}{...--multilingual-cased} & \cellcolor{ACMBlue!8}{135M} & \cellcolor{ACMBlue!8}{37.49$\pm$2.08} & \cellcolor{ACMBlue!8}{80.61$\pm$0.61} \\
\hline
deberta-v3-base & 184M & 41.97$\pm$3.12  & 82.54$\pm$0.50  \\
\cellcolor{ACMBlue!8}{mdeberta-v3-base} & \cellcolor{ACMBlue!8}{276M} & \cellcolor{ACMBlue!8}{38.18$\pm$4.40} & \cellcolor{ACMBlue!8}{81.71$\pm$0.64}  \\
\hline
roberta-base & 125M & 40.35$\pm$2.09 & 81.81$\pm$0.35 \\
    \cellcolor{ACMBlue!8}{xlm-roberta-base} & \cellcolor{ACMBlue!8}{279M} & \cellcolor{ACMBlue!8}{\textbf{41.80$\pm$2.76}} & \cellcolor{ACMBlue!8}{\textbf{81.56$\pm$0.59}} \\
\bottomrule
\end{tabular}}
\end{table}

In Table \ref{tab:multi-lingual-base-model-search}, we compare the macro and weighted F1-scores between the English and multi-lingual variants of BERT, DistilBERT, DeBERTa, and RoBERTa models. Our analysis reveals several key patterns. While all multi-lingual variants show increased parameter counts due to expanded token vocabularies for non-English languages, they remain computationally efficient with fewer than 300M parameters, resulting in negligible inference time increases. The transition to multi-lingual variants generally leads to a decrease in weighted F1-scores across all models. BERT, DistilBERT, and DeBERTa exhibit similar performance degradation patterns, with macro F1-scores dropping by approximately 3-4\%. Notably, XLM-RoBERTa demonstrates exceptional behavior, showing a slight improvement in macro F1-score compared to its English-only counterpart.

\subsubsection{Multi-lingual Dataset (\MLSNT)}

\begin{table*}[htbp!]
\centering
\caption{\MLSNT \xspace Dataset Overview. \% of lines discarded ranges from 10-70\%. In most cases, the \% of toxic lines within the dataset increases.}
\label{tab:multi-lingual-dataset-processed}

\begin{minipage}{\linewidth}
\centering
\begin{tabular}{ll|ll|lll}
\toprule
 &  & \multicolumn{2}{c|}{\textbf{Lines}} & \multicolumn{3}{c}{\textbf{Toxicity \%}} \\
\multirow{-2}{*}{\textbf{Language}} & \multirow{-2}{*}{\textbf{Name}} & \textbf{Processed} & \textbf{\% Discarded} & \textbf{Original} & \textbf{Processed} & \textbf{$\Delta$} \\ \midrule
Chinese   (Simplified) & COLD \cite{deng-etal-2022-cold} & 20,087 & 46.18\% & 48.03\% & 60.67\% & 12.64\% \\
Chinese   (Simplified) & SWSR \cite{JIANG2022100182} & 5,708 & 36.32\% & 34.50\% & 47.85\% & 13.35\% \\
Chinese   (Simplified) & TOXICN \cite{lu-etal-2023-facilitating} & 8,500 & 29.23\% & 53.79\% & 56.51\% & 2.71\% \\
Chinese   (Traditional) & TOCAB \cite{10.1109/IRI51335.2021.00069} & 65,263 & 37.25\% & 14.48\% & 8.94\% & -5.54\% \\ \hline
French & MLMA \cite{ousidhoum-etal-2019-multilingual} & 3,203 & 20.20\% & 79.55\% & 93.57\% & 14.02\% \\ \hline
German & GAHD \cite{goldzycher-etal-2024-improving} & 7,886 & 28.28\% & 42.43\% & 55.88\% & 13.45\% \\
German & GERM\_EVAL \cite{germ_eval} & 4,546 & 45.93\% & 33.76\% & 53.70\% & 19.94\% \\
German & HASOC \cite{10.1145/3368567.3368584} & 1,431 & 69.35\% & 11.63\% & 32.42\% & 20.79\% \\ \hline
Japanese & Inspection AI \footnote{\url{https://github.com/inspection-ai/japanese-toxic-dataset}} & 324 & 25.86\% & 35.93\% & 16.98\% & -18.95\% \\
Japanese & LLM\_JP \cite{DBLP:journals/corr/abs-2407-03963} & 1,662 & 10.02\% & 44.72\% & 45.43\% & 0.71\% \\ \hline
Portuguese   (Brazil) & OffCom \cite{brasnam} & 577 & 44.14\% & 19.55\% &  26.86\% & 7.31\% \\
Portuguese   (Brazil) & OLID \cite{10.1007/s10579-023-09657-0} & 5,534 & 20.40\% & 85.39\% & 94.00\% & 8.62\% \\
Portuguese   (Brazil) & ToLD \cite{leite-etal-2020-toxic} & 15,065 & 28.26\% & 44.07\% & 49.98\% & 5.91\% \\ \hline
Russian & Abusive \cite{DBLP:conf/raslan/AndrusyakRK18} & 1,184 & 40.80\% & 32.70\% & 53.97\% & 21.27\% \\
Russian & South\_Park \cite{saitov-derczynski-2021-abusive} & 13,155 & 17.13\% & 32.83\% & 32.69\% &  -0.14\% \\
\bottomrule
\end{tabular}
\end{minipage}

\end{table*}

Using prompt v1 at temperature 0.7, we prompt GPT-4o-mini across all 15 datasets. Table \ref{tab:multi-lingual-dataset-processed} shows the resulting processed dataset, where we show the amount of lines left after processing (where both human and GPT-4o-mini agreed), \% of lines that were discarded, toxicity percentage from the original and processed dataset. A key observation is the variability in the percentage of lines discarded during processing, ranging from 10.02\% (LLM\_JP) to 69.35\% (HASOC). The lines discarded do not show a clear correlation with the change in toxicity percentage. Notably, most datasets show an increase in the percentage of toxic lines after processing, particularly in languages like German (GERM\_EVAL: 19.94\% increase, HASOC: 20.79\% increase) and Russian (Abusive: 21.27\% increase). The toxicity percentages vary significantly across languages and datasets, with some, like the Russian South Park and Japanese LLM\_JP datasets, showing minimal change (-0.14\% and 0.71\% respectively), while others demonstrate substantial shifts. This variation highlights the nuances in toxicity and the importance of language-specific considerations.

\subsubsection{Multi-lingual ToxBuster}

\begin{table*}[ht!]
\centering
\caption{Comparison of mono-lingual (BERT, RoBERTa) and multi-lingual (XLM-RoBERTa) models. BERT-base achieves the best overall performance on gaming data, while the multilingual models maintain competitive performance and enable cross-lingual generalization to \MLSNT.}
\label{tab:multi-lingual-model-performance}
\begin{tabular}{cl|llll}
\toprule
\multicolumn{1}{l}{\textbf{F1-Score}} & \textbf{Model} & \textbf{GAME\_1} & \textbf{GAME\_2} & \cellcolor{ACMBlue!8}{\textbf{Overall}} & \textbf{\MLSNT} \\ \midrule
\multirow{4}{*}{Macro} & bert-base & 45.39$\pm$1.01 & 40.94$\pm$1.38 & \cellcolor{ACMBlue!8}{43.17} & \multicolumn{1}{c}{-} \\
& roberta-base & 42.85$\pm$2.11  &  38.56$\pm$2.69 & \cellcolor{ACMBlue!8}{40.71} & \multicolumn{1}{c}{-} \\
 & xlm-roberta-base & 42.78$\pm$1.75 & 38.64$\pm$1.39 & \cellcolor{ACMBlue!8}{40.71} & 42.12$\pm$1.68 \\
 & xlm-roberta-base-adapted & 42.82$\pm$1.42 & 39.09$\pm$1.59 & \cellcolor{ACMBlue!8}{40.96} & 42.51$\pm$2.71 \\ \hline
\multirow{4}{*}{Weighted} & bert-base & 82.15$\pm$0.41 & 86.81$\pm$0.44 & \cellcolor{ACMBlue!8}{84.48} & \multicolumn{1}{c}{-} \\
& roberta-base & 81.95$\pm$0.47  & 86.67$\pm$0.58
 & \cellcolor{ACMBlue!8}{84.31} & \multicolumn{1}{c}{-} \\
 & xlm-roberta-base & 81.70$\pm$0.30 & 86.42$\pm$0.29 & \cellcolor{ACMBlue!8}{84.06} & 86.44$\pm$0.33 \\
 & xlm-roberta-base-adapted & 81.79$\pm$0.51 & 86.46$\pm$0.32 & \cellcolor{ACMBlue!8}{84.13} & 86.15$\pm$0.32 \\ \bottomrule
\end{tabular}
\end{table*}

We then trained ToxBuster into its multi-lingual variant using multi-lingual base transformer models, specifically xlm-roberta-base and xm-roberta-base-adapted. We report the macro and weighted F1-score in Table \ref{tab:multi-lingual-model-performance} on our original game datasets (GAME\_1 and GAME\_2), overall performance on the game dataset, and finally on \MLSNT. 

Similar to the multi-lingual base model search results, bert-base outperforms roberta-base (43.17 -> 40.71) in terms for macro f1-score. 

By using soft-prompting to include \MLSNT, we see that going from English only to multi-lingual does not hurt the performance. Finallly, we see that the domain adaption of xlm-roberta-base, xlm-roberta-base-adapted has a slight improved scores compared to its base version

\subsubsection{Human-eval on multi-lingual game chat}

\begin{table}[htbp!]
\centering
\caption{Human evaluation of multi-lingual toxicity detection performance. German chat achieves the highest performance, while Japanese chat and other languages show significant disparities between Macro and Weighted F1-scores, indicating uneven detection capabilities across toxicity categories.}
\label{tab:multi-lingual-human-eval}

\begin{tabular}{ll|ll}
\toprule
\multicolumn{1}{c}{Language} & Game & Macro & Weighted \\
\midrule
\multirow{2}{*}{French} & GAME\_2 & 45.27\% & 82.78\% \\
 & GAME\_1 & 36.31\% & 65.83\% \\ \hline
\multirow{2}{*}{German} & GAME\_2 & 56.82\% & 75.57\% \\
 & GAME\_1 & 58.88\% & 61.09\% \\ \hline
Japanese & GAME\_1 & 19.07\% & 72.29\% \\ \hline
\multirow{2}{*}{Portuguese} & GAME\_2 & 32.96\% & 66.70\% \\
 & GAME\_1 & 48.09\% & 68.63\% \\ \hline
\multirow{2}{*}{Russian} & GAME\_2 & 45.21\% & 62.01\% \\
 & GAME\_1 & 38.83\% & 57.13\% \\ 
 \bottomrule
\end{tabular}
\end{table}

Using the fine-tuned xlm-roberta-base-adapted, we sampled inferred lines from each language and game, for a total of 450 lines per game and langauge. We report the macro and weighted class-wise F1-score in Table \ref{tab:multi-lingual-human-eval}. This manual validation expose 
the challenges in multi-lingual toxicity detection (both in transferring from \MLSNT \xspace to game-chat and lower resource compared to English). 

We see substantial performance variations, with macro F1-scores spanning from a critically low 19.07\% for Japanese GAME\_1 to a comparatively robust 58.88\% for German GAME\_1. We note that for Japanese, the chat was mostly adversarial / translated from another language and did not actually contain normal comprehensible Japanese. These disparities underscore the significant linguistic and contextual complexities inherent in cross-lingual toxicity classification. The performance heterogeneity across languages — exemplified by French's macro score of 45.27\% and weighted score of 82.78\% — illuminates challenges of developing a generalized multi-lingual toxicity detection systems. Such variations likely stem from linguistic divergences, cultural communication norms, platform-specific discourse patterns, and potentially divergent annotation strategies across different linguistic datasets.

\section{Conclusion}

In this study, we addressed two fundamental challenges in industrializing toxicity detection models for in-game chat systems: unifying game-specific models and extending multi-lingual capabilities. Our soft-prompting approach demonstrated that multiple game-specific models can be effectively unified without sacrificing performance, achieving a macro F1-score of 43.16\% while significantly reducing operational complexity and training time. The LLM-assisted label transfer framework not only enabled efficient cross-lingual toxicity classification but also provided a scalable solution for adapting toxicity definitions as organizational needs evolve. More specifically, through the integration of XLM-RoBERTa and our new \MLSNT (Multi-Lingual Social Network Toxicity) dataset spanning 15 sources across seven languages, we developed a production-ready multi-lingual ToxBuster without sacrificing it's performance on the English dataset.  Human evaluation on real game chat data revealed significant performance variations across languages (19.07\% to 58.88\% macro F1-score), where some even surpassed the English benchmark of 45.39\%.

\section{Limitations}

Our work provides significant advances in multi-game, multi-lingual toxicity detection while acknowledging several areas for future research. Our evaluation focused on two popular Ubisoft games — representing different gaming genres with distinct player interactions — though future work could explore additional gaming genres. The LLM-assisted label transfer method successfully reduces manual annotation requirements, though annotation needs vary across languages (10-70\%), suggesting opportunities for further optimization. Our human evaluation revealed expected performance variations across languages, particularly for those with limited training data or complex writing systems, highlighting the importance of continued data collection efforts. The dynamic nature of online communication presents inherent challenges, as expression patterns evolve and vary across cultural contexts. These observations align with broader industry challenges in content moderation and suggest promising directions for future research in cross-cultural toxicity detection.

\begin{acks}
We wish to thank Ubisoft La Forge,  Ubisoft Montreal User Research Lab and Ubisoft Data Office for providing technical support and insightful comments on this work. We also acknowledge funding in support of this work from Ubisoft, the Canadian Institute for Advanced Research (CIFAR AI Chair Program) and Natural Sciences and Engineering Research Council of Canada (NSERC) Postgraduate Scholarship-Doctoral (PGS D) Award.
\end{acks}


\printbibliography

\appendix
\section{Appendix}

\begin{table*}[]
\centering
\caption{Multi-lingual Human-Annotated Datasets Overview.}
\label{tab:dataset_overivew}
\begin{tabular}{l|lllll}
\toprule
\textbf{Language} & \textbf{Name} & \textbf{Task} & \textbf{Platform} & \textbf{Lines} & \textbf{Cost*} \\ \midrule
Chinese (Simplified) & COLD & Offensive & Zhihu,   Weibo, etc. (SNS) & 37,480 & \$3.64 \\
Chinese (Simplified) & SWSR & Sexism & Weibo (SNS) & 8,969 & \$0.89 \\
Chinese (Simplified) & TOXICN & Toxicity \& more & Zhihu,   Tieba   (SNS) & 12,011 & \$1.15 \\
Chinese (Traditional) & TOCAB & Profanity & PTT, a famous BBS site & 104,002 & \$9.78 \\ \hline
French & MLMA & Hate & Twitter & 4,014 & \$0.38 \\ \hline
German & GAHD & Hate & News, synthetic data & 10,996 & \$1.04 \\
German & GERM\_EVAL & Offensive & Twitter & 8,407 & \$0.81 \\
German & HASOC & Offensive, Hate & Twitter and Facebook & 4,669 & \$0.45 \\ \hline
Japanese & Inspection AI & Toxicity & -- & 437 & \$0.04 \\
Japanese & LLM\_JP & Toxicity & -- (long form text) & 1,847 & \$0.51 \\ \hline
Portuguese (Brazil) & OffCom & Offensive & Posts & 1033 & \$0.10 \\
Portuguese (Brazil) & OLID & Toxicity & Twitter, YouTube, and more & 6,952 & \$0.66 \\
Portuguese (Brazil) & ToLD & Offensive & Twitter & 21,000 & \$1.99 \\ \hline
Russian & Abusive & Abusive & Video comments & 2,000 & \$0.20 \\
Russian & South\_Park & Hate Speech & South Park (video subtitles) & 15,875 & \$1.57 \\ \bottomrule
\end{tabular}
\end{table*}

\begin{figure*}[t]
\begin{lstlisting}
You are a chat moderator for a game company. Your task is to classify whether the given line of game chat is toxic or not.

Below are the definitions for "toxic" categories.

    Definitions:
        1. "Threats (Life Threatening)" - concrete intention to cause harm, injury or other hostile actions to someone or to oneself
        2. "Minor Endangerment" -  intention or mentioning of where a child is / will be in a potentially harmful, dangerous, or inappropriate situation (e.g. grooming, neglect, pedophilia, child pornography, threats to a child, a child's personally identifiable information).
        3. "Threats (Non-Life Threatening)" - provocative statement that would cause a disruption in another player's life or game experience, without a clear intent to cause real-life injury or damage.
        4. "Hate" -  verbal abuse (e.g. intimidation, ridicule, derogatory or insulting remarks) based on another person's or group of people's actual or perceived identity (e.g., race, religion, color, sex, gender identity, national origin, age, disability, sexual orientation, genetic information).
        5. "Sexual Content / Harassment" - any text that contains, mentions or coerces other players with implicit / explicit sexual language, overtones, behavior or favors (e.g. euphemisms, flirting, innuendos, unwanted sexual advances, sexual requests)
        6. "Extremism" - political term, describes activities that are not in accordance with norms of the state; those who are fully intolerant toward others and reject the way of problem solving and existing social order.
        7. "Insults" - intent to damage the dignity of another person (or group of people), undermining their reputation or undermining their self-esteem.
        6. "Controversial / Potentially Toxic Topic" - Any text that contain any of the following subtopic(s):
            a. "Abortion": any mentions pro or against the termination of pregnancy
            b. "Religion": any text relating to / believing in a religion
            c. "Politics": any text related to how land is governed or how governments make rules and laws to manage human society.
            d. "Vulgar Content": any text that is ugly and crass (e.g. a crude joke)
            e. "Shocking / Disgusting Content": any text that contains gruesome, graphic or disgusting accounts / imagery
            f. "Hard Drugs": any mentions of strong and usually illegal drugs, likely to cause addiction
            g. "Alcohol": any mentions of alcohol or specific type of alcohol (e.g. beer, wine, vodka, ethanol, ethyl)
            h. "PII": stands for Personally Identifiable Information, any information that can be used to tie back to a person.
            i. "Trolling": deliberate intent of a player to disrupt another player's online experience.
            j. "Cheating": intent of using various methods / third party apps to create an advantage beyond normal gameplay
            k. "Scams and Advertisements": fraud, scamming (e.g. phishing), posting inappropriate links (malware, dangerous websites), advertisement of websites, services, cheats or other products.
            l. "Spamming": excessive sharing of the same / similar words, phrases or emojis, repeatedly sharing of nonsensical comments
            m. "Competitors": mention of a person, product or company related to the game at hand.
            n. "Other Offensive Content": any text that is not covered by any other class / subtopic and is offensive and / or harms a player's reasonable enjoyment of the game.

Use the following step-by-step instructions when classifying the line.

    Step 1 - The user will provide you with preceding chat lines as CONTEXT and text for the CURRENT_LINE.
    Step 2 - Familiarize yourself with the number of unique players and the lines they typed out.
    Step 3 - Based on the CONTEXT, decide if the CURRENT_LINE is "toxic" or "non-toxic". A line is toxic if at least one span belongs to one of the toxic categories. If "non-toxic", we skip step 4.
    Step 4 - Identify and classify the spans of toxic words based on the previously provided definitions. Below are some guidelines.
        a. A span can belong to one or more toxic category.
        b. If a span is classified as "Controversial / Potentially Toxic Topic", please also include the subtopic(s) the span is related to.
        c. "Threats (Life Threatening)" and "Threats (Non-Life Threatening)" are exclusive of each other.
        d. "Extremism" and ""Controversial / Potentially Toxic Topic", with the subtopic of "Politics" would appear together.

Think about it step by step.

Return in a json format where the first field is the "overall_category", being either "toxic" or "non-toxic".
If the overall_category is "toxic", return in the field "spans" a list which contains the corresponding fields {"text": "", "category: [""], }.
Example:
1. {"overall_category": "toxic"
 "spans": [
    {"text": "retard", "category": ["Insults"]}
 ]
}
2. {"overall_category": "non-toxic"}

\end{lstlisting}
\label{fig:prompt_v1}
\caption{System prompt (v1) used for label generation.}
\end{figure*}

\end{document}